\documentclass{article}
\usepackage[preprint]{neurips_2026}
\usepackage[T1]{fontenc}
\usepackage[utf8]{inputenc}
\usepackage{amsmath,amssymb,mathtools,booktabs,graphicx,microtype,xcolor}
\usepackage{array,longtable,listings,url}
\usepackage{courier}
\usepackage[colorlinks=true,linkcolor=teal!65!black,citecolor=teal!65!black,urlcolor=teal!65!black]{hyperref}
\definecolor{accent}{HTML}{087F8C}
\definecolor{soft}{HTML}{F1F6F7}
\newcommand{\E}{\mathbb{E}}
\newcommand{\Var}{\operatorname{Var}}
\newcommand{\Beta}{\operatorname{Beta}}
\newcommand{\Bern}{\operatorname{Bernoulli}}
\newcommand{\ALC}{\operatorname{ALC}}
\newcommand{\ourmethod}{Source Beta}
\title{Fast Bayesian Methods for Uncertainty Quantification\\in Sparse AI Evaluation}
\author{\textbf{Dario Deniz Ergun\textsuperscript{1} \enspace Haohua Wang\textsuperscript{1} \enspace Brendon Tao\textsuperscript{1}}\\
\textbf{Jonah Kollenberg\textsuperscript{2} \enspace Chen Ye Gan\textsuperscript{1}}\\
\textsuperscript{1}Institute of Data and Information, Tsinghua University\\
\textsuperscript{2}Delft Institute of Applied Mathematics,  Delft University of Technology
}
\begin{document}
\maketitle
\begin{abstract}
Estimating how an AI system will perform on a new benchmark need not require another large model. We use public evaluation counts to choose a weak Beta prior, then update it exactly from at most 31 revealed outcomes. The predictor uses no item semantics, pretrained weights, or inference API. Its single online evaluation in the AIMS competition returned a team-reported Brier ALC of \textbf{0.188477}, a \textbf{24.6\%} reduction from the constant-$0.5$ reference. The complete submission is \textbf{7,754 bytes}; a server microbenchmark measures \textbf{12.71~$\mu$s} per prediction with 31 matching labels. We describe how source evaluations are compressed into 301 prior means and show why a concentration of two lets new evidence rapidly determine the prediction. The resulting posterior gives exact uncertainty about the modelled success rate without numerical integration or neural inference. A public evaluation covers 79 system--benchmark pairs and eight seeds, reaching Brier ALC 0.200250 for the submitted method. Accompanying code and sufficient statistics reconstruct the fitted prior byte for byte and replay all 632 evaluation trajectories. The result demonstrates how much predictive value a compact Bayesian model can extract from sparse target feedback at very low online cost.
\end{abstract}
\section{Introduction}
Evaluating an AI system requires running it on tasks, but a large evaluation can be expensive. Before collecting many responses, we would like to estimate the probability that the system succeeds on a new benchmark. Public results give an initial expectation. A few observed outcomes then tell us how much that expectation should change. The question is how much useful prediction can be obtained from these two sources of information with a small computational cost.

The AIMS competition makes this question concrete \citep{aims}. A program receives the attributes of an evaluated system and a benchmark item, together with a small number of revealed binary outcomes. It predicts the probability of a positive response without running the evaluated system. We therefore estimate \emph{another system's behavior}. The predictor does not need to answer the question itself.

We begin with one unknown success rate for each system--benchmark pair. Public evaluation counts determine the mean of a Beta prior, while its concentration is fixed at two. This gives source knowledge the weight of only two observations: enough to specify a starting probability, but little enough for the target outcomes to replace it quickly. Beta--Bernoulli conjugacy then gives the prediction directly,
\begin{equation}
 \widehat p_n=\frac{2m+s}{2+n},\label{eq:prediction}
\end{equation}
where $m$ is the source prior mean, $n$ is the number of matching acquired labels, and $s$ is their number of successes.

The calculation uses a prior mean and two counts. It needs no representation of the item text and no numerical posterior approximation. In our single online evaluation, submission \texttt{1039683} returned Brier ALC 0.188477. Relative to the constant-$0.5$ reference, this is a 24.6\% reduction in probability error, obtained with a 7.6~KiB program and microsecond-scale measured prediction time. The small runtime information state should not be confused with a small source collection: fitting the prior uses millions of permitted public responses.

We will describe how source counts determine the prior, derive how observed labels change it, and then test what this simple construction gains and loses. The contributions are:
\begin{itemize}
\setlength{\itemsep}{1pt}\setlength{\parskip}{0pt}
\item A complete empirical-Bayes competition predictor with a benchmark-balanced source mean and exact adaptation from sparse target feedback.
\item An explicit account of prior sensitivity, posterior uncertainty, and the computational work required by the tested implementation.
\item A reproducible evaluation and prior-sensitivity study over 79 public pairs and eight seeds, separating the single online result from controlled development evidence.
\end{itemize}
Estimating a prior from related observations follows the empirical-Bayes principle \citep{robbins1956}; conjugacy itself is established methodology. Our contribution is the implementation and analysis. The analysis isolates the influence of the prior while keeping the submitted update and observed evidence fixed.

\section{Prediction task and evaluation}
Let $x=(u,i)$ contain a complete supplied system configuration $u$ and an item $i$. A pair is identified by that full configuration and the item's anonymous benchmark identifier. Different harnesses, reasoning settings, or other configuration fields define different evidence groups, even when the model name is the same. The target is the recorded binary outcome $Y\in\{0,1\}$ under the evaluation harness. Its semantics are inherited from the source evaluation; in some tasks a positive event measures attack success rather than desirable behavior.

At budgets $\mathcal B=\{0,1,3,7,15,31\}$, the predictor receives nested sets of acquired outcomes $D_n$. Acquisition and evaluation use disjoint items, and repeated responses for an item remain together. All evaluation responses contribute to the Brier loss $(\widehat p-Y)^2$ \citep{brier1950}. The normalized area under the curve on the $\log_2(n+1)$ axis is
\begin{equation}
 \ALC=0.1B_0+0.2B_1+0.2B_3+0.2B_7+0.2B_{15}+0.1B_{31},\label{eq:alc}
\end{equation}
computed within each pair and then averaged equally across pairs. Smaller values are better. Constant prediction $0.5$ gives $0.25$ at every budget.

Brier loss is a proper scoring rule \citep{gneiting2007}. If the true conditional positive probability is $p_*$, then
\begin{equation}
 \E[(q-Y)^2\mid x,D_n]=p_*(1-p_*)+(q-p_*)^2.\label{eq:brier}
\end{equation}
Thus probability estimation, rather than drawing a binary prediction, is the relevant decision. Under a Bayesian model, the posterior predictive mean minimizes the corresponding posterior expected loss.

\section{A weak source prior with exact adaptation}
\subsection{Estimating the prior from public evaluations}
For each eligible source system--benchmark record $r$, let $S_r,F_r$ denote the counts of positive and negative responses. We first form a smoothed rate
\begin{equation}
 q_r=\frac{S_r+1/2}{S_r+F_r+1}.
\end{equation}
The half-count smoothing keeps finite records away from the boundary. It is a source-rate estimator, separate from the concentration used for adaptation on a new benchmark.

For a subset of source records $\mathcal R$, let $c_b(\mathcal R)$ be the number of its records from benchmark $b$. We estimate
\begin{equation}
 m(\mathcal R)=\frac{\sum_{r\in\mathcal R}q_r/c_{b(r)}(\mathcal R)}
 {\sum_{r\in\mathcal R}1/c_{b(r)}(\mathcal R)}.\label{eq:source}
\end{equation}
Each represented benchmark therefore contributes equally, regardless of how many source configurations it contains. This avoids allowing a large evaluation table to determine the prior merely through its size. Counts within a record still use all eligible repeated outcomes; no claim of independent trials is required to compute the statistic.

We retain source records with at least 20 distinct items. A model name is normalized by trimming whitespace and applying Unicode case folding. Its matched subset is used only when it contains at least two source families; otherwise the predictor uses the global eligible subset. The exported asset contains 301 supported model identities and global mean $m_0=0.5067429803$. The final export uses all eligible source records. In offline evaluation, the complete target family is excluded before computing either the model-specific or fallback mean.

\subsection{Posterior prediction and prior sensitivity}
For a target pair, the working model is
\begin{equation}
 \theta\sim\Beta(\kappa m,\kappa(1-m)),\qquad
 Y_j\mid\theta\sim\Bern(\theta),\qquad\kappa=2.\label{eq:model}
\end{equation}
After $s$ successes and $f$ failures, conjugacy gives
\begin{equation}
 \theta\mid D_n\sim\Beta(a_n,b_n),\qquad
 a_n=2m+s,\quad b_n=2(1-m)+f,\label{eq:posterior}
\end{equation}
with $n=s+f$. Equation~\eqref{eq:prediction} is its predictive mean. No numerical integration, optimization, or MCMC is needed to implement this update.

For $n>0$, the same expression is a convex combination,
\begin{equation}
 \widehat p_n=\underbrace{\frac{2}{n+2}}_{\text{prior weight}}m
 +\underbrace{\frac{n}{n+2}}_{\text{evidence weight}}\frac{s}{n}.\label{eq:shrinkage}
\end{equation}
At 31 labels, only $2/33\simeq6.1\%$ of the weight remains on the prior. More generally, changing the source mean from $m$ to $m'$ changes the prediction by exactly
\begin{equation}
 |\widehat p_n(m)-\widehat p_n(m')|=\frac{2}{n+2}|m-m'|.\label{eq:sensitivity}
\end{equation}
This is an algebraic robustness property with respect to the prior mean. It does not protect against a nonrepresentative acquired sample or an inadequate pair-level model. Concentration two was a candidate in our development sweep; it was not inferred from a posterior over concentration values.

\begin{figure}[t]
 \centering\includegraphics[width=\linewidth]{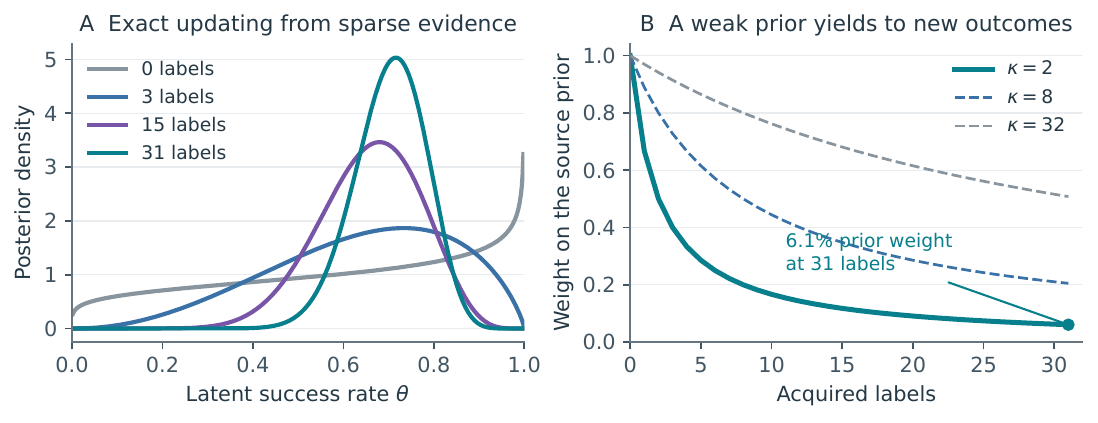}
 \caption{\textbf{How the selected model learns.} Left: illustrative Beta densities with $m=0.6$ and $(n,s)=(0,0),(3,2),(15,10),(31,22)$. These are synthetic counts explaining the mechanism, not benchmark results. Right: the exact prior weight $\kappa/(\kappa+n)$; the selected $\kappa=2$ yields rapidly to observations. All plotted coordinates are supplied as CSV files.}
 \label{fig:posterior}
\end{figure}

\subsection{What the uncertainty means}
The posterior describes uncertainty about a \emph{pair-level rate}. Writing $\mu_n=\widehat p_n$, its variance and the next-outcome variance are
\begin{align}
 \Var(\theta\mid D_n)&=\frac{\mu_n(1-\mu_n)}{n+3},\label{eq:epistemic}\\
 \Var(Y_*\mid D_n)&=\underbrace{\E[\theta(1-\theta)\mid D_n]}_{\text{outcome variability}}
 +\underbrace{\Var(\theta\mid D_n)}_{\text{uncertainty about the rate}}
 =\mu_n(1-\mu_n).\label{eq:decomp}
\end{align}
The epistemic term is bounded by $1/[4(n+3)]$, but it need not decrease after every realized label: an unexpected outcome can move the posterior mean toward $1/2$. The expected posterior variance decreases under the working model. Equal-tailed intervals for $\theta$ can be computed from Beta quantiles, although the submission returns only the predictive mean required by the interface.

These uncertainty statements are conditional on the exchangeable Bernoulli model and a fitted source mean. Item heterogeneity, dependence among outcomes, and uncertainty in the estimated source prior are not fully represented. In particular, a narrow posterior interval is not a validated guarantee of item-level calibration or real-world coverage.

\subsection{Implementation and acquisition}
The runtime loads a small JSON prior table once and reconstructs counts from the evidence supplied to each call. Exact equality of the complete subject dictionary and benchmark identifier prevents information from unrelated pairs from entering the update. Reconstruction also makes rollback to a smaller budget safe: no later labels remain in mutable posterior state. The core computation is summarized below; the bundled submission preserves the tested implementation exactly.
\begin{lstlisting}[language=Python]
name = str(subject.get("normalized_name", "")).strip().casefold()
m = model_means.get(name, global_mean)
matched = [(x, y) for x, y in labeled
           if x[0] == subject
           and x[1]["benchmark_id"] == item["benchmark_id"]]
s = sum(y for _, y in matched)
return float((2 * m + s) / (2 + len(matched)))
\end{lstlisting}
The archive omits the optional acquisition file and uses the evaluator's random policy. Within this model, every item in a pair has the same latent rate, so information-based utilities cannot distinguish candidates without introducing additional structure. Random acquisition also matches the offline protocol used to evaluate the submitted predictor.

\section{Data and experimental design}
All competition-specific fitting uses the permitted public \texttt{measurement-db} release \citep{measurementdb}, pinned to revision \texttt{54074453da7b9911be0f48e4c560be547bc84d21}. We processed 86 source wrappers under a conservative quarantine policy. After binary-outcome, condition-ambiguity, and exact-text filtering, the intermediate bank contains 11,330 pairs and 17,694,172 responses. Applying the selected model's 20-item threshold leaves \textbf{11,251 records from 84 wrappers and 83 families}, containing 17,693,497 responses. Appendix~\ref{app:data} lists every wrapper; the package contains the exact aggregate counts used for fitting.

Source construction excludes nonempty item texts that exactly match the public development panel, recording 48,178 source-item exclusions. This is a sum across wrappers, not a count of unique original questions. Known related benchmark families are quarantined, and every offline prior additionally excludes its target family. These safeguards do not certify independence of paraphrases, translations, shared original question banks, or foundation-model pretraining.

The public development panel contains 79 system--benchmark pairs, 42 benchmarks, and 19,520 recorded responses. We use eight seeds, $11,23,47,59,71,101,137,181$, with the same acquired label stream across predictors for a given pair and seed. Each split assigns half the distinct items to acquisition and half to evaluation; repeated outcomes stay with their item. We average over seeds within a pair and then equally across pairs. All six budgets are evaluated, giving 632 pair--seed trajectories per method. This panel has been repeatedly inspected and tuned against; it is an exploratory development set.

For \ourmethod, 46 of the 79 pairs use the global fallback after target-family exclusion. Sparse identity coverage therefore limits how often the method actually exploits a model-specific mean. 

\section{Results: learning from sparse feedback}
\subsection{The online submission}
The team developed and tested the predictor locally, then evaluated the selected archive once online as Codabench submission \texttt{1039683}. That run returned a team-reported Brier ALC of \textbf{0.188477}. A predictor that always returns 0.5 has Brier ALC 0.25 on any binary-outcome sample. The submitted model therefore reduces this reference error by \textbf{0.061523}, or \textbf{24.6\%}. It obtains this result by looking up a prior mean and counting matching labels, without interpreting the item text or running a neural model.

This online result concerns the all-source exported prior. The public evaluation below excludes each target family when fitting its prior. The two scores describe their respective evaluation settings; they are not a before--after comparison. The original detailed online scoring log is not bundled.

\subsection{Adaptation and calibration}
\begin{figure}[t]
\centering\includegraphics[width=\linewidth]{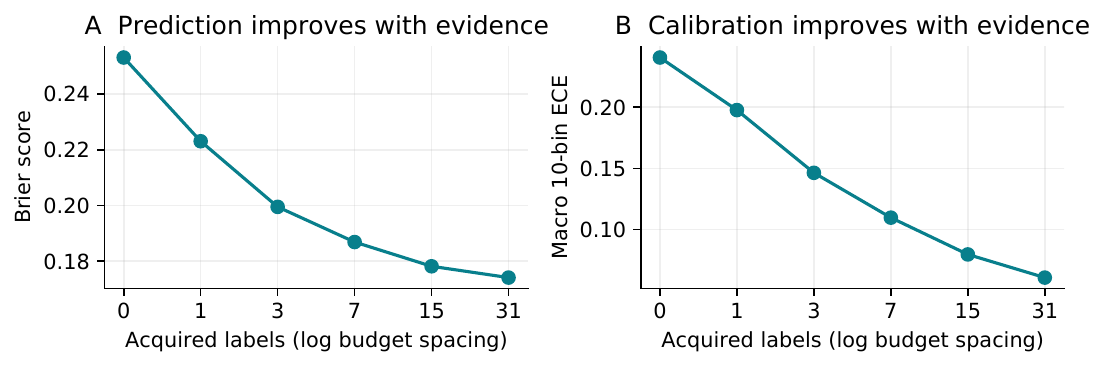}
\caption{\textbf{The submitted predictor learns from target feedback.} Brier score and macro calibration error are evaluated on the same 79 public pairs and eight seeds. Both improve as outcomes replace the initial prior expectation. These curves are public development results; the single online score is reported separately.}
\label{fig:learning}
\end{figure}

On the public panel, the submitted update reaches Brier ALC \textbf{0.200250}. Its Brier score falls from 0.253063 at zero labels to \textbf{0.174150} at 31 labels. The progression is visible in Figure~\ref{fig:learning}: each additional budget gives the target evidence more influence over the starting expectation. At 31 labels, the source prior contributes only 6.1\% of the posterior mean.

The calibration diagnostics improve at the same time. Macro 10-bin expected calibration error decreases from 0.240556 to \textbf{0.060802}, and negative log likelihood decreases from 0.699249 to \textbf{0.518072}. These diagnostics answer a complementary question: whether the assigned probabilities agree with observed response frequencies. Appendix~\ref{app:calibration} supplies the full per-budget table and reliability curves.

\section{How much does the prior matter?}
\subsection{Exact resistance to the initial mean}
The update makes its prior sensitivity explicit. If two priors have the same concentration $\kappa$ and means $m,m'$, then after the same $n$ labels,
\begin{equation}
 |\widehat p_n(m)-\widehat p_n(m')|=\frac{\kappa}{\kappa+n}|m-m'|.
\end{equation}
With the submitted $\kappa=2$, the multiplier falls from one at zero labels to $2/33$ at 31 labels. Thus \textbf{93.9\% of the initial difference is removed}. Even changing the starting mean from 0.1 to 0.9 changes the 31-label prediction by only $0.8(2/33)=0.048485$. The property follows from the update itself; it does not require that the source mean be close to the target rate.

The same calculation explains why the concentration matters. A large concentration preserves a source assumption for longer. A weak concentration lets observed outcomes determine the prediction sooner. We chose two pseudo-observations so that source knowledge supplies a starting point while target evidence rapidly takes over.

\subsection{Replaying alternative priors on identical evidence}
We test this mechanism directly by replaying the supplied public sufficient statistics. We change only the prior mean or concentration; the acquired labels, evaluation outcomes, budgets, pairs, and seeds stay fixed. Constant means $0.1,0.25,0.5,0.75,0.9$ test widely separated starting beliefs at concentration two. A second sweep keeps the fitted source means and varies concentration over $0.5,1,2,8,32$. These are sensitivity settings of the same Beta predictor, not different prediction methods.

\begin{figure}[t]
\centering\includegraphics[width=\linewidth]{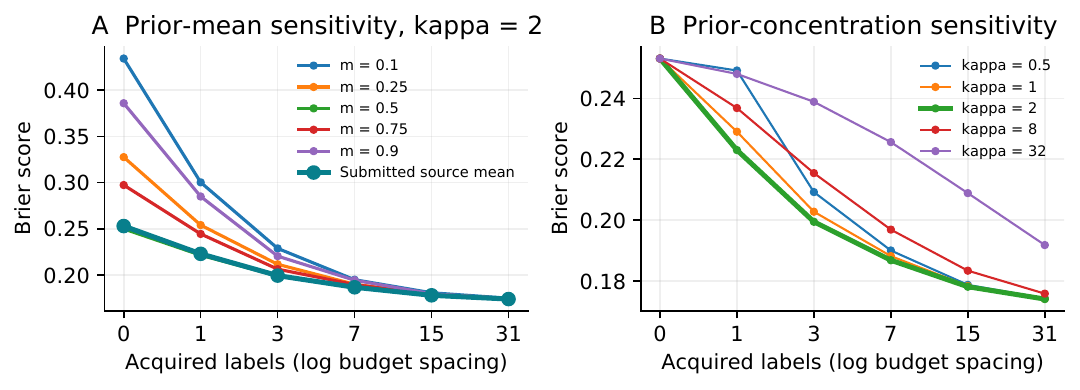}
\caption{\textbf{Target evidence reduces sensitivity to prior choice.} Left: alternative prior means with the submitted concentration two, alongside the fitted source mean. Right: the source means with alternative concentrations. Every curve replays the same public evidence. Weak priors converge rapidly toward predictions determined by the acquired outcomes.}
\label{fig:priors}
\end{figure}

Table~\ref{tab:priors} gives the exact replay scores. Among the tested concentrations with the same source mean, the submitted $\kappa=2$ has the lowest ALC. Across the source mean and constant means from 0.1 to 0.9, the 31-label Brier scores occupy the narrow range \textbf{0.174150--0.175078}, despite widely different starting beliefs. At concentration two, the uniform mean 0.5 gives ALC 0.199570 and $B_{31}=0.174163$, compared with 0.200250 and 0.174150 for the source mean. Their final-budget difference is about $1.3\times10^{-5}$. The source mean does not improve average ALC on this panel; the robust behavior comes from revising the weak starting assumption rapidly. Early budgets still contribute to ALC, so resistance at 31 labels does not make the initial prior irrelevant to the integrated score.

\begin{table}[t]\centering\small
\caption{\textbf{Prior sensitivity within the submitted Beta update.} Exact replay of 632 public pair--seed trajectories. Source means are fitted with the target family excluded. All settings use the same outcomes. Lower scores are better.}
\label{tab:priors}\begin{tabular}{lrrr}\toprule
Prior mean & Concentration & Brier ALC & $B_{31}$\\\midrule
Source mean & 0.5 & 0.208198 & 0.174364\\
Source mean & 1 & 0.202443 & 0.174250\\
Source mean & 2 & 0.200250 & 0.174150\\
Source mean & 8 & 0.209440 & 0.175928\\
Source mean & 32 & 0.228781 & 0.191871\\
$0.1$ & 2 & 0.241971 & 0.175078\\
$0.25$ & 2 & 0.217299 & 0.174597\\
$0.5$ & 2 & 0.199570 & 0.174163\\
$0.75$ & 2 & 0.211078 & 0.174189\\
$0.9$ & 2 & 0.232017 & 0.174424\\
\bottomrule\end{tabular}

\end{table}

\section{Microsecond prediction and evidence scaling}
The submitted predictor performs no transformer forward pass, matrix multiplication, or GPU computation. It scans the supplied evidence, counts matching outcomes, and evaluates Equation~\eqref{eq:prediction}. With $M$ supplied records and $n\leq31$ matches, the implementation uses $O(M+n)$ time and $O(n)$ temporary storage, plus the fixed prior dictionary. Exact configuration matching prevents unrelated evidence from entering the posterior. Reconstructing counts at each call also permits returning to a smaller label budget safely.

\begin{figure}[t]\centering
\includegraphics[width=\linewidth]{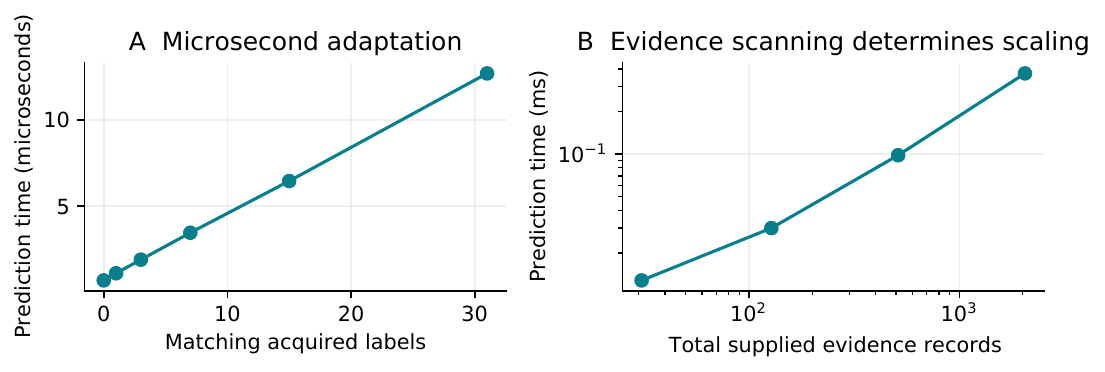}
\caption{\textbf{Runtime of the submitted predictor.} Left: measured warm prediction time as matching evidence grows. Right: packet-size stress with 31 matches and additional unrelated records. Measurements use the frozen CPU implementation.}
\label{fig:runtime}
\end{figure}

We measured the frozen implementation on a shared server with two Intel Xeon Platinum 8358P CPUs. The process was restricted to two logical cores. Each budget uses nine blocks of 1,000 predictions; quoted medians summarize block-average per-call times. At 31 matching labels, the median is \textbf{12.71~$\mu$s}. With 2,048 supplied records but still 31 matches, it is \textbf{0.372~ms}. Figure~\ref{fig:runtime} shows that the evidence scan determines scaling.

The complete tested archive is \textbf{7,754 bytes}. Fresh-process readiness has a median of 25.05~ms over three recorded trials. Fitting the prior table from loaded aggregate counts took 15.32~ms in one recorded server run. These timings exclude the earlier source-data curation. The selected prior uses millions of public source responses; its compact deployment state results from aggregating them into a small table.

For $Q$ predictions at each budget, only-matching packets require $57Q$ record inspections across all six budgets because $0+1+3+7+15+31=57$. The implementation performs scalar arithmetic, string comparisons, and memory accesses, with zero transformer or dense-matrix multiply--accumulates. The measured 31-label workload therefore captures a small and explicit computation.

\section{Scope and limitations}
The predictor estimates a pair-level rate and does not distinguish item difficulty. Its posterior uncertainty is conditional on the exchangeable Bernoulli model and fitted source mean. The public panel was reused during development, and interval coverage is not evaluated. The single online run gives no repeated-sample uncertainty estimate. These boundaries motivate testing the same weak-prior update on fresh target collections.

\section{Conclusion}
A prior mean and two counts are enough to construct a useful predictive evaluator. Our single online run reaches Brier ALC 0.188477, 24.6\% below the constant-$0.5$ reference, using a 7.6~KiB archive and 12.71~$\mu$s measured prediction time at 31 labels. A weak prior turns public knowledge into a starting expectation and then yields to target evidence: after 31 labels, only 6.1\% of its initial influence remains. Exact sensitivity formulas and matched-evidence replays make this behavior visible. The result shows how much predictive value an explicit Bayesian model can obtain from sparse feedback with very little online computation.

\paragraph{Data, code, and assistant disclosure.}
The companion release supplies the tested submission, prior-fitting and replay code, public aggregate data, figure inputs, and LaTeX sources. All competition-specific data originate from the permitted public release. The predictor uses no pretrained model or inference API. Codex and local Qwen models through OpenCode assisted with implementation, experiment supervision, analysis, and drafting. The authors remain responsible for the claims. Selected Codabench ID: \texttt{1039683}. Code and permitted derived statistics: \url{https://github.com/theNewRootUser/aims-fast-bayesian-evaluation}

\label{maintextend}
\clearpage
\bibliographystyle{plainnat}
\bibliography{references}
\clearpage
\appendix
\section{Posterior uncertainty and expected information value}
The Beta variance formula follows from $\E[\theta]=a/(a+b)$ and $\E[\theta^2]=a(a+1)/[(a+b)(a+b+1)]$. With $t=a+b=n+2$, it yields
\[
 \Var(\theta\mid D_n)=\frac{ab}{t^2(t+1)},\qquad
 \E[\theta(1-\theta)\mid D_n]=\frac{ab}{t(t+1)}.
\]
These two terms sum to the predictive Bernoulli variance. Conditional on the model, averaging over the next outcome gives
\begin{equation}
 \E_{Y_{n+1}\mid D_n}[\Var(\theta\mid D_n,Y_{n+1})]
 =\frac{t}{t+1}\Var(\theta\mid D_n).
\end{equation}
The expected contraction is exact, although realized posterior variance can increase. This distinction matters when describing uncertainty as ``decreasing with more information.''

For distinct future outcomes that are independent conditional on $\theta$, the expected reduction in optimal predictive Brier risk after acquiring one additional label is
\begin{equation}
 \Var_{Y_{n+1}\mid D_n}\!\left(\E[Y_*\mid D_n,Y_{n+1}]\right)
 =\frac{\mu_n(1-\mu_n)}{(n+3)^2}\geq0.
\end{equation}
All candidate items have this same value in the selected exchangeable model. A more informative item-selection rule requires a model in which candidates differ. \section{Additional calibration and computational details}\label{app:calibration}

\begin{figure}[ht]
 \centering\includegraphics[width=\linewidth]{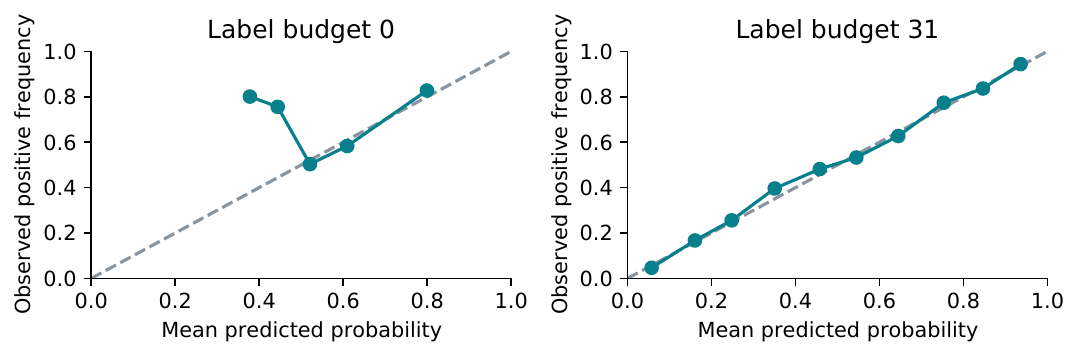}
 \caption{\textbf{Pooled reliability on the public development panel.} Bins have equal width, with contributions weighted to preserve equal pair and seed weighting. Empty bins are omitted. These pooled curves differ from averaging ECE within each pair and can conceal opposing pair-level errors. They are descriptive diagnostics and do not establish posterior-interval coverage.}
\end{figure}
\begin{table}[ht]
\caption{Selected source Beta: complete offline diagnostic curve.}
\centering\small
\begin{tabular}{rrrr}\toprule
Labels & Brier & Negative log likelihood & Macro ECE$_{10}$\\\midrule
0 & 0.253063 & 0.699249 & 0.240556 \\
1 & 0.223037 & 0.638391 & 0.197620 \\
3 & 0.199509 & 0.586371 & 0.146380 \\
7 & 0.186894 & 0.553265 & 0.109758 \\
15 & 0.178202 & 0.529858 & 0.079713 \\
31 & 0.174150 & 0.518072 & 0.060802 \\
\bottomrule\end{tabular}
\end{table}

Negative log likelihood clips predictions to $[10^{-12},1-10^{-12}]$. ECE uses ten equal-width bins within each pair and seed before averaging. The package retains both these metrics and the pooled bin masses so that aggregation choices are explicit.

The selected method's 12.71~$\mu$s timing is the median of nine block means, each computed over 1,000 calls. The reported upper percentiles in the raw evidence are percentiles of block means, not individual-call tail latencies. Separate process-startup trials use a minimal standard-library harness and warm operating-system file caches. They exclude first-time internet setup and the original data curation. A Linux \texttt{/proc/self/status} audit of a fresh selected-model process, with a 31-label synthetic fixture, records 12,197,888 resident/high-water bytes (11.63~MiB). This isolated measurement is more informative for deployment than the earlier harness-level \texttt{getrusage} high-water statistic. It is one fixture measurement, not a certified memory limit.

\section{Data disclosure and source inventory}\label{app:data}
The source snapshot is the permitted \texttt{aims-foundations/measurement-db} release. The selected predictor uses only aggregate success/failure counts and supplied model identities; it does not use private Codabench responses, web retrieval, external inference APIs, or pretrained weights. Missing and nonbinary outcomes are excluded. Source item--subject combinations with ambiguous observed test conditions are excluded. Retained repeated outcomes contribute to aggregate rates. The full processed bank and the final eligible bank are distinct stages, as shown below.

\begin{center}\small
\begin{tabular}{lrr}\toprule
 Stage & Source pairs & Binary responses\\\midrule
 Curated intermediate bank & 11,330 & 17,694,172\\
 At least 20 distinct items per pair & 11,251 & 17,693,497\\\bottomrule
\end{tabular}
\end{center}

The processed inventory contains 86 wrappers. Of these, 84 contribute eligible records to the selected prior. Table~\ref{tab:sources} lists all processed wrappers; an asterisk marks one with no eligible record. The conservative family map merges selected related wrappers, leaving 83 eligible families. The exact counts, configuration, exclusion manifest, and policy files are supplied with the report. The final asset is reconstructed from the compressed aggregate-count file, whose decompressed SHA256 is
\begin{center}\footnotesize\ttfamily
c4bc1ce814669e829212df29f56062026b031c5f101a8c76d645aea6b7260eaf
\end{center}
The frozen public development panel is identified by SHA256
\begin{center}\footnotesize\ttfamily
ffe4849e7a665fff08e0d9affb9ebb7a385c29ff68a755c0c9e8cb882e40f832.
\end{center}
The package does not redistribute raw task texts. It supplies the public aggregate counts actually used to fit the selected prior, together with 3,792 pair--seed--budget sufficient-statistic rows for exact score replay. For each such row, the prediction and evaluation mean suffice because the constant-prediction Brier loss equals $p^2-2p\bar y+\bar y$. These are public development aggregates, not hidden online test data. All plotted experimental coordinates derive from included evidence; the mechanism illustration is explicitly synthetic.

\begin{longtable}{lrlr}
\caption{Processed source wrappers and eligible record counts. Asterisks mark zero eligible records.}\label{tab:sources}\\
\toprule Wrapper & Records & Wrapper & Records\\\midrule\endfirsthead
\toprule Wrapper & Records & Wrapper & Records\\\midrule\endhead
\bottomrule\endfoot
\texttt{abstract\_reason} & 1 & \texttt{judgetuning} & 7\\
\texttt{adaptivestep} & 2 & \texttt{kernelbench} & 10\\
\texttt{aegis} & 1 & \texttt{legal\_rag\_bench} & 6\\
\texttt{afrieval} & 948 & \texttt{legalbench} & 31\\
\texttt{afrimedqa} & 32 & \texttt{lingoly} & 11\\
\texttt{ai2d\_test} & 254 & \texttt{liveaopsbench} & 13\\
\texttt{algotune} & 18 & \texttt{llm\_bp\_tag} & 1\\
\texttt{alignment\_faking} & 4278 & \texttt{llmail\_inject} & 22\\
\texttt{alpha\_sql} & 1 & \texttt{mathvista\_mini} & 263\\
\texttt{appworld} & 18 & \texttt{mind2web} & 1\\
\texttt{averimatec*} & 0 & \texttt{mj\_bench} & 55\\
\texttt{babilong} & 71 & \texttt{mmbench\_v11} & 251\\
\texttt{beavertails} & 4 & \texttt{mtbbench} & 25\\
\texttt{benger} & 29 & \texttt{naturebench} & 23\\
\texttt{bertaqa} & 6 & \texttt{nester} & 3\\
\texttt{bird\_sql} & 120 & \texttt{openbiorq} & 11\\
\texttt{brace} & 28 & \texttt{os\_harm} & 7\\
\texttt{causal\_axioms} & 2 & \texttt{perfcodebench} & 20\\
\texttt{cellverse} & 1 & \texttt{perumedqa} & 11\\
\texttt{ceval} & 11 & \texttt{pxplore} & 14\\
\texttt{clasheval} & 6 & \texttt{real\_webagents} & 33\\
\texttt{crow} & 34 & \texttt{realtimeqa} & 92\\
\texttt{dave} & 8 & \texttt{reliancescope} & 24\\
\texttt{devbench} & 19 & \texttt{risebench} & 23\\
\texttt{disco} & 4 & \texttt{robust\_reasoning\_benchmark} & 31\\
\texttt{editbench} & 44 & \texttt{safeagentbench} & 6\\
\texttt{edu\_circuit\_hw} & 6 & \texttt{sgrades} & 65\\
\texttt{egoschema} & 11 & \texttt{sib200} & 3308\\
\texttt{emoji\_attack} & 10 & \texttt{sketchjudge} & 38\\
\texttt{engdesign} & 12 & \texttt{statqa} & 38\\
\texttt{enginemt\_qa*} & 0 & \texttt{subjective\_qa} & 5\\
\texttt{erbench} & 8 & \texttt{sugarcrepe} & 1\\
\texttt{exploitgym} & 4 & \texttt{sustainable\_food} & 6\\
\texttt{faithcot} & 4 & \texttt{tensortrust} & 3\\
\texttt{felm} & 1 & \texttt{truthfulqa\_mc} & 150\\
\texttt{fewshot\_ttt\_bbh} & 11 & \texttt{tulu\_human\_eval} & 4\\
\texttt{hallusionbench} & 253 & \texttt{visual\_riddles} & 8\\
\texttt{healthadminbench} & 2 & \texttt{visualwebarena} & 3\\
\texttt{helm\_thaiexam} & 42 & \texttt{wikihow\_agent} & 12\\
\texttt{hle} & 16 & \texttt{wmt\_mqm} & 84\\
\texttt{igakuqa} & 5 & \texttt{workarena} & 2\\
\texttt{igakuqa119} & 1 & \texttt{workbench\_revisited} & 24\\
\texttt{ikp} & 170 & \texttt{worldcentralbanks} & 10\\
\end{longtable}

Upstream source licenses and attribution conditions remain applicable. Only compact derived statistics and the prior asset are redistributed; the package does not relicense raw benchmark texts.

\section{Reproducing and identifying the submission}
The selected archive is \texttt{aims\_beta\_modelprior\_k2.zip}. It contains \texttt{model.py}, its standard-library runtime, the prior JSON, and the original README and manifest. Its SHA256 is
\begin{center}\footnotesize\ttfamily
681db0f9cdaea6b0ac062c63f521a5117605509fb089dfdc9d6d697b7ca0af54.
\end{center}
The original archive passed the official submission checker and actual prediction smoke \citep{baseline}. The report-building audit uses explicit compensated summation to reconstruct the prior file byte for byte from the source counts, recomputes family-held-out means, and recovers ALC 0.2002499321275406 over 632 trajectories to floating-point precision. The publication bundle has a separate file manifest; modifying the paper does not alter the tested submission ZIP.

\begin{lstlisting}[language=bash]
# Standard library only: reconstruct priors and replay offline scores.
python code/reproduce_selected.py
# Rebuild every image from the bundled evidence and plot data.
python -m pip install -r requirements-figures.txt
python code/make_selected_figures.py
# Compile the paper using the unmodified official NeurIPS style.
tectonic main.tex
# Alternatively: latexmk -pdf main.tex
\end{lstlisting}

The release separates executable reproduction code from historical development snapshots. The former is self-contained for the selected prior and score replay. The latter records the original filtering and evaluation procedures and their server paths; rebuilding the entire raw-data campaign additionally requires the pinned public release and historical manifests. The full public source collection remains on the lab server. Data used by every figure are included as both evidence JSON and small CSV tables, with PDF, SVG, and PNG outputs.

The report uses the official NeurIPS 2026 style in preprint mode, without modifying its margins or typography. AIMS permits nonanonymous reports of approximately four to eight main-text pages, with unlimited references and appendices. The companion \texttt{REQUIREMENTS.md} maps those requirements to this release. The selected Codabench submission ID is \texttt{1039683}. Code and permitted derived statistics:  No additional online model evaluation was performed while preparing the report.

\end{document}